\pdfoutput=1

\documentclass[11pt,a4paper]{article}
\usepackage[hyperref]{acl2021}
\usepackage{times}
\usepackage{latexsym}

\usepackage{microtype}

\aclfinalcopy

\renewcommand{\paragraph}[1]{\vspace*{.2\baselineskip}\newline\textbf{#1}\ \ }

\title{We went to look for meaning and all we got were these lousy representations: aspects of meaning representation for computational semantics}

\author{Simon Dobnik, Robin Cooper, Adam Ek, Bill Noble, Staffan Larsson, \\ \textbf{Nikolai Ilinykh, Vladislav Maraev and Vidya Somashekarappa}\thanks{\quad Euqual contribution.} \\
  CLASP \& FLoV, University of Gothenburg\\
\texttt{\{name.surname\}@gu.se}}

\date{}

\begin{document}
\maketitle
\begin{abstract}
  In this paper we examine different meaning representations that are
  commonly used in different natural language applications today and
  discuss their limits, both in terms of the aspects of the natural
  language meaning they are modelling and in terms of the aspects of
  the application for which they are used.
\end{abstract}

\section{Introduction}

A crucial component to produce a ``successful'' NLP system is
sufficiently expressive representations of meaning.  We consider a
sufficiently expressive meaning representation to be one that allows a
system's output to be considered acceptable to native speakers given
the task.  In this paper we present several features of meaning and
discuss how different methods of deriving meaning representations
capture these features.  This list is by no means exhaustive.  It
might be viewed as a first attempt to discuss ways of establishing a
general methodology for evaluating meaning representations and
characterising what kinds of applications they might be useful for.
The features we will discuss are:
\paragraph{compositionality} The ability to compute the meaning of
phrases on the basis of the meanings of their immediate
sub-constituents.
\paragraph{logic-based inference} The ability to derive conclusions
based on logical inference, including logical inferences based on the
semantics of logical constants such as \textbf{and}, \textbf{not} and
logical quantifiers and also consequences that follow from additional
axioms or ``meaning postulates''.
\paragraph{discourse semantics} This involves giving meaning
representations for units larger than a sentence.
\paragraph{underspecification} Underspecified meaning representations
are single representations which cover several meanings in cases where
there is systematic ambiguity.
\paragraph{model theory} Model theory deals with representing the
relationship between language and the world.
\paragraph{dialogue} Dialogue semantics is sometimes thought of as
part of discourse semantics.  However, there are many phenomena that
occur in conversations with two or more participants that do not occur
in texts.  These include fragmentary utterances, repair phenomena,
utterances split between different dialogue participants and overlap
(dialogue participants speaking at the same time).  This seems to
warrant dialogue being treated as a feature separate from discourse.
\paragraph{similarity of meaning} In addition to meaning relations
such as entailment there is a notion of words, phrases and sentences
having similar meanings in various respects.
\paragraph{robust non-logical inference} This type of inference is
discussed in e.g. the work on textual entailment.  Rather than
representing something that follows logically, it corresponds to what
conclusions people might draw from a given utterance or text, is often
reliant on background knowledge and is to a large extent defeasible.
\paragraph{dynamic changes of meaning} The meaning of words and
phrases can change over time and during the course of a text or
dialogue.
\paragraph{grounding meaning in action and perception} While model
theory purports to relate language and the world it tells us little
about how we relate our perception of the world and action in the
world to the meaning of words and phrases.  Such issues become
important, for example, if we want to put natural language on board a
robot.
\paragraph{multimodality} The multimodal nature of communication
becomes obvious when you begin to think of meaning in terms of action
and perception.

The rigour of the work on semantics by Richard Montague
\citep{Montague1973,Partee1976} inspired early work on computational
semantics \citep[perhaps the earliest
was][]{FriedmanWarren1978,FriedmanMoranWarren1978}.  Two high-points
of the literature on computational semantics based on Montague are
\citet{BlackburnBos2005}, using logic programming, and
\citet{EijckUnger2010}, using functional programming. Montague's
semantic techniques have also played an important role in semantic
treatments using Combinatory Categorial Grammar
\citep[CCG,][]{BosClarkSteedmanCurranHockenmaier2004}.

One problem with Montague's treatment of semantics was that it was
limited to the level of the sentence.  It could not, for example, deal
with cross-sentence anaphora such as \textit{A dog$_i$ barked.  It$_i$
  was upset by the intruder}.  This, among several other things, led
to the development of Discourse Representation Theory
\citep[DRT,][]{KampReyle1993,KampGenabithReyle2011} and other variants
of \textit{dynamic} semantics such as \citet{Heim1982} and
\citet{GroenendijkStokhof1991}. Here ``dynamic'' is meant in the sense
of treating semantic content as context change potential in order,
among other things, to be able to pass referents from one sentence to
a subsequent sentence in the discourse.  This is a much less radical
notion of dynamic interpretation than we discuss in
Section~\ref{sec:dyn}, where the meaning associated with a word or
phrase may change as a dialogue progresses.  DRT has played an
important role in computational semantics from early work on the
Verbmobil project \citep{BosGambaeckLieskeMoriPinkalWorm1996} to work
by Johan Bos and others on the Groningen Meaning Bank
(\url{http://gmb.let.rug.nl/}) and the Parallel Meaning Bank
(\url{https://pmb.let.rug.nl/}).

What do we get from this body of work? Here are some of the features
that we can get in a compositional semantics based on this work.
\paragraph{compositionality} Compositionality is one of the
cornerstones of Montague's approach.
\paragraph{logic-based inference} The ability to derive conclusions
based on logical inference and the ability to characterise ``meaning
postulates'' is a central feature of semantics in the Montague
tradition.  Defeasible reasoning has been added to this kind of
framework \citep[e.g.,][]{AsherLascarides2003} and systems have been
connected to theorem provers and model builders
\citep{BlackburnBos2005}.
\paragraph{discourse semantics} The variants of dynamic semantics
discussed above gave us the ability to treat discourse phenomena (that
is, phenomena occurring in texts or utterances of more than a single
sentence, including cases of discourse anaphora).
\paragraph{underspecification} While there is some work on
underspecification of meaning in the theoretical literature
\citep{Reyle1993}, the most interest has been devoted to it in
computational work based on formal semantics \citep[such
as][]{Alshawi1992,Bos1996,CopestakeFlickingerPollardSag2005}.
\paragraph{model theory} Model theory associated with a formal
approach to semantics can in computational terms relate to database
query \citep{BlackburnBos2005,EijckUnger2010}.

What we have sketched above might be called the classical canon of
formal semantics as it relates to computational semantics.  There is
much that we would like to have for a computational semantics that is
still lacking here.  To some extent more recent developments address
these gaps and to some extent they are addressed by other kinds of
meaning representations we discuss later in the paper, though often at
the expense of giving up on (or at least having difficulty with) the
features that we listed above.  Features lacking in the classical
canon include:
\paragraph{dialogue} The notion that language is actually used in
interaction between agents engaging in communication came quite late
to formal semantics though there is now a significant body of
theoretical work such as \citet{Ginzburg1994,Ginzburg2012}.  This gave
rise to the Information State Update approach to dialogue systems
\citep{Larsson2002}.  TTR \citep[a type theory with
records,][]{Cooper2005a,Cooperinprepa} has played an important role in
this.
\paragraph{similarity of meaning} The kind of meaning similarity that
is discussed in connection with vector semantics (see
Section~\ref{sec:distrib}) is hard to recreate in a formal meaning
representation, though the use of record types in TTR suggests that a
connection could be made.
\paragraph{robust non-logical inference} The kind of inference that is
represented, for example, in work on textual entailment is hard to
square with the logic-based inference discussed above.  However, the
work on topoi by \citet{Breitholtz2020}, perhaps coupled with
probabilistic TTR \citep{CooperDobnikLappinLarsson2015}, is suggestive
of a computational approach to this.
\paragraph{dynamic changes of meaning} Notions of meaning negotiation
and coordination have become central in the literature on formal
approaches to dialogue.  We discuss this in Section~\ref{sec:dyn}.
\paragraph{grounding meaning in action and perception} This has become
central to theories such as TTR and Dynamic Syntax
\citep{KempsonCannGregoromichelakiChatzikyriakidis2016} and we discuss
this in Section~\ref{sec:ground}.
\paragraph{multimodality} The multimodal nature of communication
becomes obvious when you begin to think of meaning in terms of action
and perception.  We discuss this in Section~\ref{sec:body}.

Above we have mentioned examples of formal approaches which attempt to
incorporate features which are not present in the classical canon.  An
alternative strategy is to try to incorporate features from the
classical canon in non-formal approaches \citep[for
example,][]{CoeckeSadrzadehClark2010} or to combine aspects of
non-formal and formal approaches in a single framework \citep[for
example,][]{Larsson2013,ErkHerbelot2020}.

\section{Distributional meaning representations}
\label{sec:distrib}

Distributional representations of meaning can be traced back to
\citet{wittgenstein1953philosophical}, but was popularised by
\citet{Firth_Papers57}.  The idea at its core is that the meaning of a
word is given by its context.  \citet{wittgenstein1953philosophical}
primarily speaks about meaning in relation to the world, or some game,
while \citet{Firth_Papers57} speaks about language in relation to
language.  The second notion of meaning, is the basis for
distributional semantics.  This notion is both realised in the world
(for example if someone say ``grab" sufficiently often when grabbing
things), and in language where if two words occur is the same context,
e.g. $a$ and $b$ occur in the context of ``colour", we know that they
relate to colour.

The two predominant approaches to constructing distributional meaning
representations today is using deep learning to construct distributed
word representations and contextualised word representations
\citep{chrupala2019correlating}.  In these approaches, the meaning of
a word is encoded as a dense vector of real valued numbers. The values
of the vector is determined by the task used for training and the
corpus used to train.

Distributed word representations focus on building static
representations of words given a corpus. Popular techniques for
obtaining these representations are BoW (Bag-of-Words) or SGNS (Skip
Gram with Negative Sampling), popularised by
\citep{DBLP:journals/corr/abs-1301-3781}.  Distributed representations
construct a representation of a word $x$ such that given a context $C$
we are able to select $x$ from a vocabulary $V$.  The results of such
models is a matrix of size $(N,D)$ where the row $n_i$ is the meaning
representation of word $x_i \in V$.  Contextualised representations on
the other hand attempt to build a model which gives a word different
representations given its context (i.e. dynamic representations).  The
main difference between the two methods of meaning representations is
that in contextualised methods the matrix $(N,D)$ is accompanied by a
model $f$, typically a language model, which yields word
representations given a \textit{sentence}.

With distributed representations we may also reason analogically about
words and about combinations of concepts, e.g. "Russia" + "River" =
"Volga" \citep{DBLP:conf/nips/MikolovSCCD13}. That is, we may
construct complex meaning by combining simpler parts.  By adding the
representation for ``Russia" and ``river" we obtain some vector $z$
which contains information about the contexts of both ``Russia" and
``river". By querying the vector space for words with a
\textit{similar} representation to that of $z$ we find other words
with similar context, where we find rivers in Russia and among them,
Volga. %
The success of distributed meaning representations, both static and
contextualised, can in part be attributed to the ability of a model to
predict \textit{similarity} between units of language.  Because
meaning is defined as the context in which language occurs, two vector
representations can be compared and their similarity
measured. \citep{DBLP:conf/lrec/ConneauK18,
  DBLP:journals/corr/abs-2003-04866}.  This similarity, in the case of
words, indicate whether they occur in similar context.  Or, in the
case of sentences, indicate whether they have similar sentence
meaning. For example, the sentences (1) \textit{``the price is 1
  dollar"} and (2) \textit{``the amount to pay is 100 cents"} are
essentially equivalent.  If we consider the sentence \textit{``the
  price is 2 dollars"} it's arguably more similar to (1) and (2) than
\textit{``the cat sat on the mat"}.  This problem has been explored in
the STS (Semantic Textual Similarity) shared tasks
\citep{DBLP:journals/corr/abs-1708-00055}.

The ability to model similarity allows models to discover
relationships between units of language.  It allows models to transfer
knowledge between languages. For example, unsupervised word
translation can be done by aligning monolingual vector spaces.
\citep{DBLP:conf/iclr/LampleCRDJ18, DBLP:conf/acl/AgirreLA18}.  The
transformer models \citep{DBLP:conf/nips/VaswaniSPUJGKP17} have also
enabled zero-shot and transfer learning, e.g. by learning English word
representations and evaluating on a task in another language
\citep{pires-etal-2019-multilingual}.  The simplicity of static and
contextualised meaning representations allows us to construct meaning
representations for \textit{any} unit of language, be it words,
sentences \citep{DBLP:conf/acl/BaroniBLKC18}, documents
\citep{DBLP:conf/rep4nlp/LauB16} or languages
\citep{ostling-tiedemann-2017-continuous}.

But a word, or a sentence may mean different things depending on the
context.  For example a sentence in different domains will express
different meanings even if the words are exactly the same.  This
presents a problem for distributed representations, as our observation
of a word or sentence in the real world may differ from what we have
seen in the data.  However, the effects of different domains may be
counteracted by \textit{domain adaptation} techniques
\citep{DBLP:conf/acl/JiangZ07}.  The same holds for words, where a
word may mean different things depending on the sentence it occurs
in. This is a problem for static embeddings where a word is associated
with one and only one meaning representation. The problem is mitigated
in contextualised representations that construct the meaning
representations based on the context. That is, ``mouse" will have a
different representation when ``animal" or ``computer" is in the
sentence. This does not solve the problem but goes some way in
disentangling the different meanings.

Distributed representations enjoy success across a wide variety of NLP
tasks. However, a consequence of automatically learning from a corpus
results in some inherent shortcomings. A corpus is a snapshot of a
subset of language and only captures language as it was used then and
there. This means that the meaning representations do not model
language as it changes (see Section \ref{sec:dyn}).  Additionally, the
meaning representations are created from observing language use, not
from language use in the world.  A consequence of this is that
distributional meaning representations don't capture the
state-of-affairs in the world, i.e. the context, in which the language
was used.  In practical terms this means that for tasks that depends
on the state-of-affairs in the world, such as robot control, dialogue
or image captioning, a system needs to gather this information from
elsewhere.

\section{Dynamic meaning representations}
\label{sec:dyn}

Meaning is context dependent in (at least) two different ways.  To see
how, we can make the distinction between \emph{meaning potential} and
\emph{situated meaning} \citep{Noren2007}.  The situated meaning of a
word is its disambiguated interpretation in a particular context of
use.  Meaning potential (or \emph{lexical meaning}) is the system of
affordances \citep{Gibson1966,Gregoromichelaki2020} that a word offers
for sense-making in a given context.  In this conception, situated
meaning is context dependent by construction, but we also claim that
the meaning potential of a word depends on context of a certain kind.
In particular, it depends on what is \emph{common ground}
\citep{Stalnaker2002} between a speaker and their audience.

At a linguistics conference, a speaker might use words like
\emph{attention} or \emph{modality}--- words that would mean something
completely different (or nothing at all) at a family dinner.  The
conference speaker expects to be understood \emph{based on} their and
their audience's joint membership in the computational linguistics
community, where they (rightly or wrongly) consider certain
specialised meanings to be common ground.  The communities that serve
as a basis for semantic common ground can be as broad as
\emph{speakers of Spanish} (grounding the ``standard'' Spanish
lexicon), or as small as a nuclear family or close group of friends
(grounding specialised meanings particular to that group of people)
\citep{Clark1996}.

Recent work in NLP has demonstrated the value of modelling context,
including sentential (Section~\ref{sec:distrib}) and multimodal
context (Section~\ref{sec:ground}).  in the representation of situated
meanings.  Very little work has been done to model the \emph{social
  context}, which provides the basis for semantic common ground.  As a
result, NLP models typically assume that words have a single, fixed
lexical meaning.  We identify three related sources of lexical
fluidity that might be accounted for with dynamic meaning
representations by incorporating social context of different kinds.
\paragraph{Variation} As demonstrated in the conference example,
lexical meaning is community dependent.  This doesn't necessarily mean
that every NLP application needs to mimic the human ability to tailor
our semantic representations to the different communities we belong
to, but some applications may serve a broader set of users by doing
so.  Consider, for example, an application that serves both the
general public and experts in some domain.

Even where variation is not explicitly modelled, it is an important
factor to consider on a meta level.  In practice, NLP models typically
target the most prestigious, hegemonic dialect of a given language,
due in part to biases in what training data is easily available on the
internet \citep{Bender2021}.  This results in applications that favour
users who are more comfortable with the dominant language variety.

Furthermore, many applications \emph{assume} a single variety of a
given language, when in fact the training data of the models they rely
on is rather specific.  The standard English BERT model, for example,
is trained on a corpus of unpublished romance novels and encyclopedia
articles, but is applied as if it represents English written large.
\paragraph{Alignment} Semantic common ground is not only based on
joint community membership--- it can also be built up between
particular agents through interaction.  Additions or modifications to
existing common ground can take place implicitly (through
\emph{semantic accommodation}) or explicitly, as in \emph{word meaning
  negotiation} \citep{Myrendal2015}.  Experiments have shown that
pairs of speakers to develop shorter lexicalised referring expressions
in when they need to repeatedly identify a referent \citep{Mills2008}.

There is some hope for developing models that dynamically update their
meaning representations based on interaction with other agents.
\citet{Larsson2017} suggest an inventory of semantic update functions
that could be applied to formal meaning representations based on the
results of an explicit word meaning negotiation.  On the
distributional side, one- or few-shot learning may eventually allow
models to generalise from a small number of novel uses by drawing on
existing structure in the lexicon \citep{Lake2019}.  One question that
remains unexplored in both these cases is which updates to local
(dialogue or partner-specific) semantic ground should be propagated to
the agent's representation of the communal common ground (and to which
community).  This naturally bring up the issue of community-level
semantic change.
\paragraph{Change} How words change in meaning has long been an object
of study for historical linguists \citep[e.g.,][]{Paul1891,Blank1999}.
Historical change may not seem like a particularly important thing for
NLP applications to model.  After all, we can accommodate for changes
over decades or centuries by simply retraining models with more
current data, but significant \emph{semantic shift} can also take
place over a much shorter timeline, especially in smaller speech
communities \citep{Eckert1992}.  The issue of semantic change also
intersects with that of variation, since coinages and shifts in
meaning that take place in one community can cause the lexical common
ground to diverge from another community.  Conversely, variants in one
community may come to be adopted by another (possibly broader)
community.

The recent widespread use of distributional semantics to study
semantic change suggests that distributional representations are
capable of capturing change.%
\footnote{See \citet{Tahmasebi2018}, \citet{Tang2018}, and
  \citet{Kutuzov2018} for recent surveys.}  Diachronic distributional
representations have been used to study semantic change on both a
historic/language level \citep[e.g.,][]{Dubossarsky2015,Hamilton2016a}
and on a short-term/community level
\citep{Rosenfeld2018,DelTredici2019}.

While social context is not often taken into account in meaning
representations, ongoing research on semantic variation and change
suggests that such dynamic representations are possible as extensions
of the formal and distributional paradigms.

\section{Grounded meaning representations in action and perception}
\label{sec:ground}

The meaning of words is not merely in our head.  It is grounded in our
surroundings and tied to our understanding of the world
\cite{regier1996}, particularly through visual perception
\cite{mooney2008}. Mapping language and vision to get
\textbf{multi-modal} meaning representations imposes an important
challenge for many real-world NLP applications, e.g. conversational
agents.  This section describes how different modalities are typically
integrated to get a meaning representation for
\textbf{language-and-vision (L\&V)} tasks and what is still missing in
the respective \textbf{information fusion} techniques.

Historically, modelling of situated language has been influenced by
ideas from language technology, computer vision and robotics, where a
combination of top-down rule-based language systems was connected with
Bayesian models or other kinds of classifiers of action and perception
\cite{Kruijff:2007,Dobnik:2009dz,Tellex:2011wf,Mitchell:2012aa}.  In
these approaches, most of the focus was on how to ground words or
phrases in representations of perception and action through
classification.  Another reason for this hybrid approach has also been
that such models are partially interpretable.  Therefore, they has
been a preferred choice in critical robotic applications where
security is an issue.  The compositionality of semantic
representations in these systems is ensured by using semantic
grammars, while perceptual representations such as SLAM maps
\cite{Dissanayake:2001} or detected visual features \cite{Lowe:1999aa}
provide a model for interpreting linguistic semantic representations.
Deep learning, where linguistic and perceptual features are learned in
an interdependent manner rather than engineered, has proven to be
greatly helpful for the task of image captioning
\cite{vinyals2015tell,anderson2018bottomup,bernardi2017automatic} and
referring expression generation \cite{kazemzadeh2014}.

A more in-depth analysis of how meaning is represented in these models
is required.  \citet{Ghanimifard:2017ab} show that a neural language
model can learn compositionality by grounding an element in the
spatial phrase in some perceptual representation.  In terms of
methodology for understanding what type of meaning is captured by the
model, attention \cite{xu2016show,lu2017knowing} has been successfully
used. \newcite{lu2017hierarchical} have shown that co-attending to
image and question leads to a better understanding of the regions and
words the model is focused on the most.  \newcite{ilinykh2020}
demonstrate that attention can struggle with fusing multi-modal
information into a single meaning representation based on the human
evaluation of generated image paragraphs.  This is because the nature
of visual and linguistic features and the model's structure
significantly impact what representations can be learned when using
attention mechanism.  Examining attention shows that attention can
correctly attend to objects, but once it is tasked to generate
relations (such as prepositional spatial relations and verbs),
attention visually disappears as these relations are non-identifiable
in the visual features utilised by the model.  This leads several
researchers to include specifically geometric information in image
captioning models \cite{Sadeghi:2015aa,Ramisa:2015aa}. On the other
hand, it has also been shown that a lot of meaning can be extracted
solely from word distributions.  \citet{Choi:2020aa} demonstrates how
linguistic descriptions encode common-sense knowledge which can be
applied to several tasks while \citet{Dobnik:2013aa,Dobnik:2018ab}
demonstrate that word distributions are an important contributing part
of the semantics of spatial relations.

Interactive set-ups such as visual question answering (VQA)
\cite{agrawal2016vqa,devries2017guesswhat} or visual dialogue
\cite{das2017visual} make first attempts in modelling multi-modal
meaning in multi-turn interaction. However, such set-ups are
asymmetric in terms of each interlocutor's roles, which leads to
homogeneous question-answer pairs with rigid word meaning.
\textit{Conversational games} have been proposed as set-ups in which
the meaning of utterances is agreed upon in a collaborative setting.
These settings allow for modelling meaning coordination and phenomena
such as clarification requests \cite{schlangen2019grounded}.
\newcite{ilinykh2019meetup} propose a two-player coordination game,
MeetUp!, which imposes demands on a conversational agent to utilise
dialogue discourse and visual information to achieve a mutual
understanding with their partner.  \newcite{haber2019photobook}
introduce the PhotoBook task, in which the agent is required to be
able to track changes in the meaning of referring expressions, which
is continually changing throughout the dialogue.

Examining L\&V models and representations they learn points to several
signficiant and interesting challenges. The first relates to the
structure of both datasets and models.  Many corpora contain
prototypical scenes where the model can primarily optimise on the
information from the language model to generate an answer without even
looking at the image \cite{Cadene:2019vb}. Secondly, information
captured by a language model is more compact and expressive than
patterns of visual and geometric features.  Thirdly, common-sense
language model and visual information is not enough
\cite{Lake:2017ab,Bisk:2020aa,Tenenbaum:2020aa}: we also rely on
mental simulation of the scene's physics to estimate, for example,
from the appearance and position of a person's body that they are
making a jump on their skateboard rather than they are falling over a
fire hydrant.  Such representations are necessary for modelling
\textit{embodied agents}
\cite{anderson2018visionandlanguage,embodiedqa,kottur2018visual}.
Fourthly, adding more modalities and representations puts new
requirements on inference procedures and more sophisticated models of
attention \cite{Lavie:2004aa} that weighs to what degree such features
are relevant in a particular context.  In recent years we have seen
work along these lines implemented with a transformer architecture
\cite{lu2019vilbert,Su2019,herdade2020image}.  However, the issue of
interpretability in terms of how individual parts (self-attentions) of
large-scale models process information from different modalities is
still an open question.

\section{Representations of meaning expressed with our body}
\label{sec:body}

In this section we attempt to raise awareness of the role of our
bodies in meaning creation and perception in a bidirectional way. This
includes how meanings can result in bodily reactions and, conversely,
how meanings can be expressed with our bodies, including non-verbal
vocalisations, gaze and gestures.

\subsection{Emotions}
\label{sec:emotions}

Our view of emotions is two-fold. On one hand, meanings perceived from
the environment can change our emotional states and be expressed in
bodily reactions. For instance, evaluating events as intrinsically
unpleasant may result in gaze aversion, pupillary constriction and
some of the other components listed by \citet{scherer2009dynamic}. On
the other hand, our emotional states can be expressed and the
expressions can be adjusted by emotional components, such as mood
\citep{marsella2010computational}.

Over the last years \emph{appraisal theories} became the leading
theories of emotions \citep[for overview, see][]{oatley14}. These
theories posit that emotion arises from a person’s interpretation of
their relationship with the environment. The key idea behind cognitive
theories is that emotions not only reflect physical states of the
agents but also emotions are judgements, depending on the current
state of the affairs (depending on a certain person,
significance/urgency of the event etc.). Such an evaluation is called
\emph{appraisal}. In our view, linguistic events can as well enter the
calculation of appraisal on the level of information-state of the
agent and the formal theories of emotions can be implemented to model
this process. For instance, following the view of \citet{oatley14} we
can distinguish emotions as either free-floating or requiring an
object, whereas in the latter case the object can be a linguistic
entity, entity in the environment or a part of agent's
information-state (e.g., obstruction of the agent's goal can lead to
anger or irritation, and, vice versa, agent's sadness can lead to the
search for a new plan).

\subsection{Non-verbal vocalisations}
\label{sec:nvss}

Non-verbal vocalisations, such as laughter, are ubiquitous in our
everyday interactions. In the British National Corpus laughter is a
quite frequent signal regardless of gender and age---the spoken
dialogue part of the British National Corpus contains approximately
one laughter event every 14 utterances.  In the Switchboard Dialogue
Act corpus non-verbally vocalised dialogue acts (whole utterances
marked as non-verbal) constitute 1.7\% of all dialogue acts and
laughter tokens make up 0.5\% of all the tokens that occur in the
corpus.

A much debated question is to what extent laughter is under voluntary
control. Despite a very particular bodily reaction (laughter causes
tensions and relaxations of our bodies), it is believed that we laugh
in a very different sense from sneezing or coughing
\citep{prusak2006science}. Many scholars agree that we laugh for a
reason, \emph{about} something.  One of the most prominent arguments
against involuntary laughter is its social function, that is
well-documented \citep[e.g., ][]{mehu2011smiling}: laughter is
associated with senses of closeness and affiliation, establishing
social bonding and smoothing away discomfort.  Even the ``primitive''
case of tickling not only requires the presence of the other
(self-ticking is much less efficient), but also tickling stimulation
is likely to elicit laughter if subjects have close relationships
\citep{harris1999mystery}. Therefore, it is hard to claim that the
behaviour that is highly socially dependent can be involuntary.

This leads us to the conclusion that the meaning of laughter ought to
be represented, which would allow an artificial agent to understand it
and react accordingly
\citep{Maraev.Mazzocconi.Howes.Ginzburg_ISCA_2018}. \citet{mazzocconi2019phd}
presents a function-based taxonomy of laughter, distinguishing, for
example, such functions as indication of pleasant incongruity or
smoothing the discomfort in conversation. \citet{ginzburg2020laughter}
propose an account for formal treatment of laughter within the
information-state of dialogue participants, which includes potential
scaling up to other non-verbal social signals, namely, smiling,
sighing, eye rolling and frowning.

\subsection{Gaze}

Gaze is one of the non-verbal signals with many functions. It can
dictate attention, intentions, and serve as a communicative cue in
interaction.  Gaze following can infer the object people are looking
at. While scanning a visual scene, the brain stores the fixation
sequences in memory and reactivated it when visualising it later in
the absence of any perceptual information \cite{Brandt97}. Scan-path
theory illustrations indicate the meaning representations on areas
scanned depended on the semantics of a sentence
\cite{Bochynska2015}. The existence of semantic eye fixations supports
the view of mental imagery that is intrinsically flexible and
creative. Although it is grounded on particular previous experiences,
by selecting the past episode it is able to generalise the past
information to novel images that share features with the novel item
\cite{MartarelliFred17}.  The spatial representations associated with
semantic category launch eye movements during retrieval
\cite{Spivey00}.

For dialogue participants gaze patterns act as resources to track
their stances. Interlocutors engage in mutual gaze while producing
agreeing assessments \cite{Haddington06}.  Gaze shifts sequentially
follow a problematic stance and are followed by a divergent stance by
the person who produced the gaze shift. Gaze patterns are not
meaningful themselves but become so in dialogue, when combined with
linguistic and other interactional practices.

Eye movement patterns, EEG signals and brain imaging are some of the
techniques that have been widely used to augment traditional
text-based features. Temporal course and flexibility with respect to
speakers eye gaze can be used to disambiguate referring expressions in
spontaneous dialogue. Eye tracking data from reading provide
structural signal with fine-grained temporal resolution which closely
follows the sequential structure of speech and is highly related with
the cognitive workload associated with speech processing
\cite{Barrett20}.

Also, CNN has been used to learn features from both gaze and text to
classify the input text yielding better classification performance by
leveraging the eye movements obtained from human readers to tackle
semantic challenges better \cite{Mishra18}. For multimodal and
multiparty interaction in both social and referential scenarios,
\citet{Vidya20} calls for categorical representation of gaze patterns.

\subsection{Gestures}

Gestures are the hand and body movements that help convey information
\cite{Kita03}. The observational, experimental, behavioural and
neuro-cognitive evidence indicate that language and gestures are
tightly linked in comprehension and production \cite{David06,
  Willems07}.  Speech and gestures are semantically and temporally
coordinated and therefore involved in co-production of meaning.

Meanings are conveyed by gestures through iconicity and spacial
proximity providing information that are not necessarily expressed in
speech (e.g., size and shape). Even though the shaping of gestures is
related to the conceptual and semantic aspects of accompanying speech,
gestures cannot be unambiguously interpreted by naïve listeners
\cite{Hadar04}. While \citet{Morett20}, showed that the semantic
relationship between representational gestures and their lexical
affiliates are evaluated similarly regardless of language modality.

The mentions of referents for the first time in discourse are often
accompanied by gestures. \citet{Debreslioska} report that ``entity''
gesture accompanies referents expressed by indefinite nominals.  The
clause structures specialise for the introduction of the referents,
which contrasts the representation of ``action'' gestures that
co-occur with inferable referents expressed by definite nominals.

Fixing gesture functions, integrating the representations originating
from different modalities and determining their composite meanings is
challenging.  To develop an agent system, multimodal output planning
is crucial and timing should be explicitly
represented. \citet{lucking2016} approaches some of the challenges
from type-theoretic perspective, representing iconic gestures in TTR
\citep{Cooper2005a} and linking them with linguistic predicates.

\section{Conclusions}

We surveyed formal, distributional, interactive, multi-modal and
body-related representations of meaning used in computational
semantics.  Overall, we conclude, they are able to deal with
compositionality, under-specification, similarity of meaning,
inference and provide an interpretation of expressions but in very
different ways, capturing very different kinds of meaning.  While this
works well in practice for individual systems, a challenge arises when
we try to combine representations.  What do joint representations
represent?  How they can be transferred across-contexts of language
use?

In line with this we suggest that future work should focus on
developing benchmarks that compare and test these representations. %
We hope that this paper points to some of the aspects of
representations that need to be taken into account.

\section*{Acknowledgments}

The research reported in this paper was supported by a grant from the Swedish Research Council (VR project 2014-39) for the establishment of the Centre for Linguistic Theory and Studies in Probability (CLASP) at the University of Gothenburg.

\bibliography{references}

\begin{thebibliography}{127}
\expandafter\ifx\csname natexlab\endcsname\relax\def\natexlab#1{#1}\fi

\bibitem[{Alshawi(1992)}]{Alshawi1992}
Hiyan Alshawi, editor. 1992.
\newblock \emph{{The Core Language Engine}}.
\newblock MIT Press.

\bibitem[{{Anderson} et~al.(2018{\natexlab{a}}){Anderson}, {He}, {Buehler},
  {Teney}, {Johnson}, {Gould}, and {Zhang}}]{anderson2018bottomup}
P.~{Anderson}, X.~{He}, C.~{Buehler}, D.~{Teney}, M.~{Johnson}, S.~{Gould}, and
  L.~{Zhang}. 2018{\natexlab{a}}.
\newblock \href {https://doi.org/10.1109/CVPR.2018.00636} {Bottom-up and
  top-down attention for image captioning and visual question answering}.
\newblock In \emph{2018 IEEE/CVF Conference on Computer Vision and Pattern
  Recognition}, pages 6077--6086.

\bibitem[{{Anderson} et~al.(2018{\natexlab{b}}){Anderson}, {Wu}, {Teney},
  {Bruce}, {Johnson}, {Sünderhauf}, {Reid}, {Gould}, and {van den
  Hengel}}]{anderson2018visionandlanguage}
P.~{Anderson}, Q.~{Wu}, D.~{Teney}, J.~{Bruce}, M.~{Johnson}, N.~{Sünderhauf},
  I.~{Reid}, S.~{Gould}, and A.~{van den Hengel}. 2018{\natexlab{b}}.
\newblock \href {https://doi.org/10.1109/CVPR.2018.00387} {Vision-and-language
  navigation: Interpreting visually-grounded navigation instructions in real
  environments}.
\newblock In \emph{2018 IEEE/CVF Conference on Computer Vision and Pattern
  Recognition}, pages 3674--3683.

\bibitem[{Antol et~al.(2015)Antol, Agrawal, Lu, Mitchell, Batra, Zitnick, and
  Parikh}]{agrawal2016vqa}
Stanislaw Antol, Aishwarya Agrawal, Jiasen Lu, Margaret Mitchell, Dhruv Batra,
  C.~Lawrence Zitnick, and Devi Parikh. 2015.
\newblock {VQA}: {V}isual {Q}uestion {A}nswering.
\newblock In \emph{International Conference on Computer Vision (ICCV)}.

\bibitem[{Artetxe et~al.(2018)Artetxe, Labaka, and
  Agirre}]{DBLP:conf/acl/AgirreLA18}
Mikel Artetxe, Gorka Labaka, and Eneko Agirre. 2018.
\newblock \href {https://doi.org/10.18653/v1/P18-1073} {A robust self-learning
  method for fully unsupervised cross-lingual mappings of word embeddings}.
\newblock In \emph{Proceedings of the 56th Annual Meeting of the Association
  for Computational Linguistics, {ACL} 2018, Melbourne, Australia, July 15-20,
  2018, Volume 1: Long Papers}, pages 789--798. Association for Computational
  Linguistics.

\bibitem[{Asher and Lascarides(2003)}]{AsherLascarides2003}
N~Asher and A~Lascarides. 2003.
\newblock \emph{Logics of conversation}.
\newblock Cambridge University Press.

\bibitem[{Barrett and Hollenstein(2020)}]{Barrett20}
Maria Barrett and Nora Hollenstein. 2020.
\newblock \href {https://doi.org/10.1111/lnc3.12396} {Sequence labelling and
  sequence classification with gaze: Novel uses of eye‐tracking data for
  natural language processing}.
\newblock \emph{Language and Linguistics Compass}, 14.

\bibitem[{Bender et~al.(2021)Bender, Gebru, {McMillan-Major}, and
  Shmitchell}]{Bender2021}
Emily~M. Bender, Timnit Gebru, Angelina {McMillan-Major}, and Shmargaret
  Shmitchell. 2021.
\newblock \href {https://doi.org/10.1145/3442188.3445922} {On the {{Dangers}}
  of {{Stochastic Parrots}}: {{Can Language Models Be Too Big}}? \&\#x1f99c;}.
\newblock In \emph{Proceedings of the 2021 {{ACM Conference}} on {{Fairness}},
  {{Accountability}}, and {{Transparency}}}, {{FAccT}} '21, pages 610--623,
  {New York, NY, USA}. {Association for Computing Machinery}.

\bibitem[{Bernardi et~al.(2016)Bernardi, Cakici, Elliott, Erdem, Erdem,
  Ikizler-Cinbis, Keller, Muscat, and Plank}]{bernardi2017automatic}
Raffaella Bernardi, Ruket Cakici, Desmond Elliott, Aykut Erdem, Erkut Erdem,
  Nazli Ikizler-Cinbis, Frank Keller, Adrian Muscat, and Barbara Plank. 2016.
\newblock Automatic description generation from images: A survey of models,
  datasets, and evaluation measures.
\newblock \emph{J. Artif. Int. Res.}, 55(1):409–442.

\bibitem[{Bisk et~al.(2020)Bisk, Holtzman, Thomason, Andreas, Bengio, Chai,
  Lapata, Lazaridou, May, Nisnevich, Pinto, and Turian}]{Bisk:2020aa}
Yonatan Bisk, Ari Holtzman, Jesse Thomason, Jacob Andreas, Yoshua Bengio, Joyce
  Chai, Mirella Lapata, Angeliki Lazaridou, Jonathan May, Aleksandr Nisnevich,
  Nicolas Pinto, and Joseph Turian. 2020.
\newblock \href {https://arxiv.org/abs/2004.10151} {Experience grounds
  language}.
\newblock \emph{arXiv}, arXiv:2004.10151 [cs.CL].

\bibitem[{Blackburn and Bos(2005)}]{BlackburnBos2005}
Patrick Blackburn and Johan Bos. 2005.
\newblock \emph{Representation and Inference for Natural Language: A First
  Course in Computational Semantics}.
\newblock CSLI Studies in Computational Linguistics. CSLI Publications,
  Stanford.

\bibitem[{Blank(1999)}]{Blank1999}
A.~Christian Blank. 1999.
\newblock Why do new meanings occur? {{A}} cognitive typology of the
  motivations for lexical semantic change.
\newblock In Andreas Blank and Peter Koch, editors, \emph{Historical
  {{Semantics}} and {{Cognition}}}. {De Gruyter Mouton}.

\bibitem[{Bochynska and Laeng(2015)}]{Bochynska2015}
Agata Bochynska and Bruno Laeng. 2015.
\newblock \href {https://doi.org/10.1007/s10339-015-0690-0} {Tracking down the
  path of memory: eye scanpaths facilitate retrieval of visuospatial
  information}.
\newblock \emph{Cognitive processing}, 16 Suppl 1.

\bibitem[{Bos(1996)}]{Bos1996}
J.~Bos. 1996.
\newblock Predicate logic unplugged.
\newblock In \emph{{Proceedings of the Tenth Amsterdam Colloquium}}, pages
  133--143, Amsterdam. {ILLC/Department of Philosophy, University of
  Amsterdam}.

\bibitem[{Bos et~al.(2004)Bos, Clark, Steedman, Curran, and
  Hockenmaier}]{BosClarkSteedmanCurranHockenmaier2004}
Johan Bos, Stephen Clark, Mark Steedman, James~R Curran, and Julia Hockenmaier.
  2004.
\newblock {Wide-coverage semantic representations from a CCG parser}.
\newblock In \emph{COLING 2004: Proceedings of the 20th International
  Conference on Computational Linguistics}, pages 1240--1246.

\bibitem[{Bos et~al.(1996)Bos, Gamb{\"a}ck, Lieske, Mori, Pinkal, and
  Worm}]{BosGambaeckLieskeMoriPinkalWorm1996}
Johan Bos, Bj{\"o}rn Gamb{\"a}ck, Christian Lieske, Yoshiki Mori, Manfred
  Pinkal, and Karsten Worm. 1996.
\newblock {Compositional semantics in Verbmobil}.
\newblock \emph{arXiv preprint cmp-lg/9607031}.

\bibitem[{Brandt and Stark(1997)}]{Brandt97}
Stephan Brandt and Lawrence Stark. 1997.
\newblock \href {https://doi.org/10.1162/jocn.1997.9.1.27} {Spontaneous eye
  movements during visual imagery reflect the content of the visual scene}.
\newblock \emph{Journal of cognitive neuroscience}, 9:27--38.

\bibitem[{Breitholtz(2020)}]{Breitholtz2020}
Ellen Breitholtz. 2020.
\newblock \emph{Enthymemes in Dialogue}.
\newblock Brill.

\bibitem[{Cadene et~al.(2019)Cadene, Dancette, Cord, and
  Parikh}]{Cadene:2019vb}
Remi Cadene, Corentin Dancette, Matthieu Cord, and Devi Parikh. 2019.
\newblock \href {https://arxiv.org/pdf/1906.10169.pdf} {{RUBi}: Reducing
  unimodal biases for visual question answering}.
\newblock In \emph{NeurIPS}, pages 841--852.

\bibitem[{Cer et~al.(2017)Cer, Diab, Agirre, Lopez{-}Gazpio, and
  Specia}]{DBLP:journals/corr/abs-1708-00055}
Daniel~M. Cer, Mona~T. Diab, Eneko Agirre, I{\~{n}}igo Lopez{-}Gazpio, and
  Lucia Specia. 2017.
\newblock \href {http://arxiv.org/abs/1708.00055} {Semeval-2017 task 1:
  Semantic textual similarity - multilingual and cross-lingual focused
  evaluation}.
\newblock \emph{CoRR}, abs/1708.00055.

\bibitem[{Choi(2020)}]{Choi:2020aa}
Yejin Choi. 2020.
\newblock \href {https://youtu.be/h2wzQKRAdA8} {Intuitive reasoning as
  (un)supervised language generation}.
\newblock Seminar, Paul G. Allen School of Computer Science and Enginneering,
  University of Washington and Allen Institute for Artificial Intelligence, MIT
  Embodied Intelligence Seminar.

\bibitem[{Chrupa{\l}a and Alishahi(2019)}]{chrupala2019correlating}
Grzegorz Chrupa{\l}a and Afra Alishahi. 2019.
\newblock Correlating neural and symbolic representations of language.
\newblock \emph{arXiv preprint arXiv:1905.06401}.

\bibitem[{Clark(1996)}]{Clark1996}
Herbert~H. Clark. 1996.
\newblock \href {https://doi.org/10.1017/CBO9780511620539} {\emph{Using
  {{Language}}}}.
\newblock {Cambridge University Press}.

\bibitem[{Coecke et~al.(2010)Coecke, Sadrzadeh, and
  Clark}]{CoeckeSadrzadehClark2010}
Bob Coecke, Mehrnoosh Sadrzadeh, and Stephen Clark. 2010.
\newblock Mathematical foundations for a compositional distributional model of
  meaning.
\newblock \emph{arXiv preprint arXiv:1003.4394}.

\bibitem[{Conneau and Kiela(2018)}]{DBLP:conf/lrec/ConneauK18}
Alexis Conneau and Douwe Kiela. 2018.
\newblock \href
  {http://www.lrec-conf.org/proceedings/lrec2018/summaries/757.html} {Senteval:
  An evaluation toolkit for universal sentence representations}.
\newblock In \emph{Proceedings of the Eleventh International Conference on
  Language Resources and Evaluation, {LREC} 2018, Miyazaki, Japan, May 7-12,
  2018}. European Language Resources Association {(ELRA)}.

\bibitem[{Conneau et~al.(2018)Conneau, Kruszewski, Lample, Barrault, and
  Baroni}]{DBLP:conf/acl/BaroniBLKC18}
Alexis Conneau, Germ{\'{a}}n Kruszewski, Guillaume Lample, Lo{\"{\i}}c
  Barrault, and Marco Baroni. 2018.
\newblock \href {https://doi.org/10.18653/v1/P18-1198} {What you can cram into
  a single {\textbackslash}{\textdollar}{\&}!{\#}* vector: Probing sentence
  embeddings for linguistic properties}.
\newblock In \emph{Proceedings of the 56th Annual Meeting of the Association
  for Computational Linguistics, {ACL} 2018, Melbourne, Australia, July 15-20,
  2018, Volume 1: Long Papers}, pages 2126--2136. Association for Computational
  Linguistics.

\bibitem[{Cooper(2005)}]{Cooper2005a}
Robin Cooper. 2005.
\newblock Records and record types in semantic theory.
\newblock \emph{Journal of Logic and Computation}, 15(2):99--112.

\bibitem[{Cooper(in prep)}]{Cooperinprepa}
Robin Cooper. in prep.
\newblock \href {https://sites.google.com/site/typetheorywithrecords/drafts}
  {{From perception to communication: An analysis of meaning and action using a
  theory of types with records (TTR)}}.
\newblock Draft of book chapters available from
  \url{https://sites.google.com/site/typetheorywithrecords/drafts}.

\bibitem[{Cooper et~al.(2015)Cooper, Dobnik, Lappin, and
  Larsson}]{CooperDobnikLappinLarsson2015}
Robin Cooper, Simon Dobnik, Shalom Lappin, and Staffan Larsson. 2015.
\newblock {Probabilistic Type Theory and Natural Language Semantics}.
\newblock \emph{{Linguistic Issues in Language Technology}}, 10(4):1--45.

\bibitem[{Copestake et~al.(2005)Copestake, Flickinger, Pollard, and
  Sag}]{CopestakeFlickingerPollardSag2005}
Ann Copestake, Dan Flickinger, Carl Pollard, and Ivan~A Sag. 2005.
\newblock {Minimal recursion semantics: An introduction}.
\newblock \emph{Research on language and computation}, 3(2):281--332.

\bibitem[{Das et~al.(2018)Das, Datta, Gkioxari, Lee, Parikh, and
  Batra}]{embodiedqa}
Abhishek Das, Samyak Datta, Georgia Gkioxari, Stefan Lee, Devi Parikh, and
  Dhruv Batra. 2018.
\newblock {E}mbodied {Q}uestion {A}nswering.
\newblock In \emph{Proceedings of the IEEE Conference on Computer Vision and
  Pattern Recognition (CVPR)}.

\bibitem[{Das et~al.(2017)Das, Kottur, Gupta, Singh, Yadav, Moura, Parikh, and
  Batra}]{das2017visual}
Abhishek Das, Satwik Kottur, Khushi Gupta, Avi Singh, Deshraj Yadav,
  Jos\'e~M.F. Moura, Devi Parikh, and Dhruv Batra. 2017.
\newblock {V}isual {D}ialog.
\newblock In \emph{Proceedings of the IEEE Conference on Computer Vision and
  Pattern Recognition (CVPR)}.

\bibitem[{Debreslioska and Gullberg(2020)}]{Debreslioska}
Sandra Debreslioska and Marianne Gullberg. 2020.
\newblock \href {https://doi.org/10.3389/fpsyg.2020.01935} {What’s new?
  gestures accompany inferable rather than brand-new referents in discourse}.
\newblock \emph{Frontiers in Psychology}, 11.

\bibitem[{Del~Tredici et~al.(2019)Del~Tredici, Fern{\'a}ndez, and
  Boleda}]{DelTredici2019}
Marco Del~Tredici, Raquel Fern{\'a}ndez, and Gemma Boleda. 2019.
\newblock Short-{{Term Meaning Shift}}: {{A Distributional Exploration}}.
\newblock In \emph{Proceedings of the 2019 {{Conference}} of the {{North
  American Chapter}} of the {{Association}} for {{Computational Linguistics}}},
  volume 1 (Long and Short Papers), pages 2069--2075, {Minneapolis, Minnesota}.
  {Association for Computational Linguistics}.

\bibitem[{Dissanayake et~al.(2001)Dissanayake, Newman, Durrant-Whyte, Clark,
  and Csorba}]{Dissanayake:2001}
M.~W. M.~G Dissanayake, P.~M. Newman, H.~F. Durrant-Whyte, S.~Clark, and
  M.~Csorba. 2001.
\newblock A solution to the simultaneous localization and map building ({SLAM})
  problem.
\newblock \emph{IEEE Transactions on Robotic and Automation}, 17(3):229--241.

\bibitem[{Dobnik(2009)}]{Dobnik:2009dz}
Simon Dobnik. 2009.
\newblock \emph{Teaching mobile robots to use spatial words}.
\newblock Ph.D. thesis, University of Oxford: Faculty of Linguistics, Philology
  and Phonetics and The Queen's College, Oxford, United Kingdom.

\bibitem[{Dobnik et~al.(2018)Dobnik, Ghanimifard, and Kelleher}]{Dobnik:2018ab}
Simon Dobnik, Mehdi Ghanimifard, and John~D. Kelleher. 2018.
\newblock \href {https://www.aclweb.org/anthology/W18-1401/} {Exploring the
  functional and geometric bias of spatial relations using neural language
  models}.
\newblock In \emph{Proceedings of the First International Workshop on Spatial
  Language Understanding {(SpLU 2018)} at {NAACL-HLT 2018}}, pages 1--11, New
  Orleans, Louisiana, USA. Association for Computational Linguistics.

\bibitem[{Dobnik and Kelleher(2013)}]{Dobnik:2013aa}
Simon Dobnik and John~D. Kelleher. 2013.
\newblock \href {http://pre2013.uvt.nl/pdf/dobnik-kelleher.pdf} {Towards an
  automatic identification of functional and geometric spatial prepositions}.
\newblock In \emph{Proceedings of PRE-CogSsci 2013: Production of referring
  expressions -- bridging the gap between cognitive and computational
  approaches to reference}, pages 1--6, Berlin, Germany.

\bibitem[{Dubossarsky et~al.(2015)Dubossarsky, Tsvetkov, Dyer, and
  Grossman}]{Dubossarsky2015}
Haim Dubossarsky, Yulia Tsvetkov, Chris Dyer, and Eitan Grossman. 2015.
\newblock A bottom up approach to category mapping and meaning change.
\newblock \emph{NetWordS 2015 Word Knowledge and Word Usage}, page~5.

\bibitem[{Eckert and {McConnell-Ginet}(1992)}]{Eckert1992}
Penelope Eckert and Sally {McConnell-Ginet}. 1992.
\newblock Communities of practice: {{Where}} language, gender, and power all
  live.
\newblock \emph{Locating Power, Proceedings of the 1992 Berkeley Women and
  Language Conference}, pages 89--99.

\bibitem[{van Eijck and Unger(2010)}]{EijckUnger2010}
Jan van Eijck and Christina Unger. 2010.
\newblock \emph{{Computational Semantics with Functional Programming}}.
\newblock Cambridge University Press.

\bibitem[{Erk and Herbelot(2020)}]{ErkHerbelot2020}
Katrin Erk and Aurelie Herbelot. 2020.
\newblock How to marry a star: probabilistic constraints for meaning in
  context.
\newblock \emph{arXiv preprint arXiv:2009.07936}.

\bibitem[{Firth(1957)}]{Firth_Papers57}
J.~R. Firth. 1957.
\newblock \emph{{P}apers in {L}inguistics, 1934-1951}.
\newblock {O}xford {U}niversity {P}ress, London.

\bibitem[{Friedman et~al.(1978)Friedman, Moran, and
  Warren}]{FriedmanMoranWarren1978}
Joyce Friedman, Douglas~B. Moran, and David~S. Warren. 1978.
\newblock {Two Papers on Semantic Interpretation in Montague Grammar}.
\newblock \emph{American Journal of Computational Linguistics}.
\newblock Microfiche 74.

\bibitem[{Friedman and Warren(1978)}]{FriedmanWarren1978}
Joyce Friedman and David~S Warren. 1978.
\newblock {A parsing method for Montague grammars}.
\newblock \emph{Linguistics and Philosophy}, 2(3):347--372.

\bibitem[{Ghanimifard and Dobnik(2017)}]{Ghanimifard:2017ab}
Mehdi Ghanimifard and Simon Dobnik. 2017.
\newblock \href {http://www.aclweb.org/anthology/W/W17/W17-6808.pdf} {Learning
  to compose spatial relations with grounded neural language models}.
\newblock In \emph{Proceedings of {IWCS 2017}: 12th International Conference on
  Computational Semantics}, pages 1--12, Montpellier, France. Association for
  Computational Linguistics.

\bibitem[{Gibson(1966)}]{Gibson1966}
James~J Gibson. 1966.
\newblock \emph{{The senses considered as perceptual systems}}.
\newblock {Mifflin}, {New York [u.a.}

\bibitem[{Ginzburg(1994)}]{Ginzburg1994}
Jonathan Ginzburg. 1994.
\newblock An update semantics for dialogue.
\newblock In \emph{Proceedings of the 1st International Workshop on
  Computational Semantics}, Tilburg University. ITK Tilburg.

\bibitem[{Ginzburg(2012)}]{Ginzburg2012}
Jonathan Ginzburg. 2012.
\newblock \emph{The Interactive Stance: Meaning for Conversation}.
\newblock Oxford University Press, Oxford.

\bibitem[{Ginzburg et~al.(2020)Ginzburg, Mazzocconi, and
  Tian}]{ginzburg2020laughter}
Jonathan Ginzburg, Chiara Mazzocconi, and Ye~Tian. 2020.
\newblock Laughter as language.
\newblock \emph{Glossa: a journal of general linguistics}, 5(1).

\bibitem[{Gregoromichelaki et~al.(2020)Gregoromichelaki, Chatzikyriakidis,
  Eshghi, Hough, Howes, Kempson, Kiaer, Purver, Sadrzadeh, and
  White}]{Gregoromichelaki2020}
Eleni Gregoromichelaki, Stergios Chatzikyriakidis, Arash Eshghi, Julian Hough,
  Christine Howes, Ruth Kempson, Jieun Kiaer, Matthew Purver, Mehrnoosh
  Sadrzadeh, and Graham White. 2020.
\newblock Affordance {{Competition}} in {{Dialogue}}: {{The Case}} of
  {{Syntactic Universals}}.
\newblock In \emph{Proceedings of the 24th {{Workshop}} on the {{Semantics}}
  and {{Pragmatics}} of {{Dialogue}} - {{Full Papers}}}.

\bibitem[{Groenendijk and Stokhof(1991)}]{GroenendijkStokhof1991}
Jeroen Groenendijk and Martin Stokhof. 1991.
\newblock Dynamic predicate logic.
\newblock \emph{Linguistics and Philosophy}, pages 39--100.

\bibitem[{Haber et~al.(2019)Haber, Baumg{\"a}rtner, Takmaz, Gelderloos, Bruni,
  and Fern{\'a}ndez}]{haber2019photobook}
Janosch Haber, Tim Baumg{\"a}rtner, Ece Takmaz, Lieke Gelderloos, Elia Bruni,
  and Raquel Fern{\'a}ndez. 2019.
\newblock \href {https://doi.org/10.18653/v1/P19-1184} {The {P}hoto{B}ook
  dataset: Building common ground through visually-grounded dialogue}.
\newblock In \emph{Proceedings of the 57th Annual Meeting of the Association
  for Computational Linguistics}, pages 1895--1910, Florence, Italy.
  Association for Computational Linguistics.

\bibitem[{Hadar and Pinchas-Zamir(2004)}]{Hadar04}
Uri Hadar and Lian Pinchas-Zamir. 2004.
\newblock \href {https://doi.org/10.1177/0261927X04263825} {The semantic
  specificity of gesture}.
\newblock \emph{Journal of Language and Social Psychology - J LANG SOC
  PSYCHOL}, 23:204--214.

\bibitem[{Haddington(2006)}]{Haddington06}
Pentti Haddington. 2006.
\newblock \href {https://doi.org/10.1515/TEXT.2006.012} {The organization of
  gaze and assessments as resources for stance taking}.
\newblock \emph{Text \& Talk - TEXT TALK}, 26:281--328.

\bibitem[{Hamilton et~al.(2016)Hamilton, Leskovec, and
  Jurafsky}]{Hamilton2016a}
William~L. Hamilton, Jure Leskovec, and Dan Jurafsky. 2016.
\newblock \href {https://doi.org/10.18653/v1/P16-1141} {Diachronic {{Word
  Embeddings Reveal Statistical Laws}} of {{Semantic Change}}}.
\newblock In \emph{Proceedings of the 54th {{Annual Meeting}} of the
  {{Association}} for {{Computational Linguistics}} ({{Volume}} 1: {{Long
  Papers}})}, pages 1489--1501, {Berlin, Germany}. {Association for
  Computational Linguistics}.

\bibitem[{Harris(1999)}]{harris1999mystery}
Christine~R Harris. 1999.
\newblock The mystery of ticklish laughter.
\newblock \emph{American Scientist}, 87(4):344.

\bibitem[{Heim(1982)}]{Heim1982}
Irene Heim. 1982.
\newblock \emph{{The Semantics of Definite and Indefinite NPs}}.
\newblock Ph.D. thesis, {University of Massachusetts at Amherst}.

\bibitem[{Herdade et~al.(2019)Herdade, Kappeler, Boakye, and
  Soares}]{herdade2020image}
Simao Herdade, Armin Kappeler, Kofi Boakye, and Joao Soares. 2019.
\newblock \href
  {https://proceedings.neurips.cc/paper/2019/file/680390c55bbd9ce416d1d69a9ab4760d-Paper.pdf}
  {Image captioning: Transforming objects into words}.
\newblock In \emph{Advances in Neural Information Processing Systems},
  volume~32. Curran Associates, Inc.

\bibitem[{Ilinykh and Dobnik(2020)}]{ilinykh2020}
Nikolai Ilinykh and Simon Dobnik. 2020.
\newblock \href {https://www.aclweb.org/anthology/2020.inlg-1.40} {When an
  image tells a story: The role of visual and semantic information for
  generating paragraph descriptions}.
\newblock In \emph{Proceedings of the 13th International Conference on Natural
  Language Generation}, pages 338--348, Dublin, Ireland. Association for
  Computational Linguistics.

\bibitem[{Ilinykh et~al.(2019)Ilinykh, Zarrie{\ss}, and
  Schlangen}]{ilinykh2019meetup}
Nikolai Ilinykh, Sina Zarrie{\ss}, and David Schlangen. 2019.
\newblock \href {http://semdial.org/anthology/Z19-Ilinykh_semdial_0006.pdf}
  {Meet up! a corpus of joint activity dialogues in a visual environment}.
\newblock In \emph{Proceedings of the 23rd Workshop on the Semantics and
  Pragmatics of Dialogue - Full Papers}, London, United Kingdom. SEMDIAL.

\bibitem[{Jiang and Zhai(2007)}]{DBLP:conf/acl/JiangZ07}
Jing Jiang and ChengXiang Zhai. 2007.
\newblock \href {https://www.aclweb.org/anthology/P07-1034/} {Instance
  weighting for domain adaptation in {NLP}}.
\newblock In \emph{{ACL} 2007, Proceedings of the 45th Annual Meeting of the
  Association for Computational Linguistics, June 23-30, 2007, Prague, Czech
  Republic}. The Association for Computational Linguistics.

\bibitem[{Kamp et~al.(2011)Kamp, van Genabith, and
  Reyle}]{KampGenabithReyle2011}
Hans Kamp, Josef van Genabith, and Uwe Reyle. 2011.
\newblock \href {https://doi.org/10.1007/978-94-007-0485-5_3} {{Discourse
  Representation Theory}}.
\newblock In Dov Gabbay and Franz Guenthner, editors, \emph{{Handbook of
  Philosophical Logic}}, volume~15. {Springer Science+Business Media B.V. }.

\bibitem[{Kamp and Reyle(1993)}]{KampReyle1993}
Hans Kamp and Uwe Reyle. 1993.
\newblock \emph{From Discourse to Logic}.
\newblock Kluwer, Dordrecht.

\bibitem[{Kazemzadeh et~al.(2014)Kazemzadeh, Ordonez, Matten, and
  Berg}]{kazemzadeh2014}
Sahar Kazemzadeh, Vicente Ordonez, Mark Matten, and Tamara~L. Berg. 2014.
\newblock Referit game: Referring to objects in photographs of natural scenes.
\newblock In \emph{EMNLP}.

\bibitem[{Kempson et~al.(2016)Kempson, Cann, Gregoromichelaki, and
  Chatzikyriakidis}]{KempsonCannGregoromichelakiChatzikyriakidis2016}
Ruth Kempson, Ronnie Cann, Eleni Gregoromichelaki, and Stergios
  Chatzikyriakidis. 2016.
\newblock {Language as Mechanisms for Interaction}.
\newblock \emph{Theoretical Linguistics}, 42(3-4):203--276.

\bibitem[{Kita and Özyürek(2003)}]{Kita03}
Sotaro Kita and Asli Özyürek. 2003.
\newblock \href {https://doi.org/10.1016/S0749-596X(02)00505-3} {What does
  cross-linguistic variation in semantic co-ordination of speech and gesture
  reveal?: Evidence of an interface representation of spatial thinking and
  speaking}.
\newblock \emph{Journal of Memory and Language}, 48:16--32.

\bibitem[{Kottur et~al.(2018)Kottur, Moura, Parikh, Batra, and
  Rohrbach}]{kottur2018visual}
Satwik Kottur, José M.~F. Moura, Devi Parikh, Dhruv Batra, and Marcus
  Rohrbach. 2018.
\newblock \href {http://arxiv.org/abs/1809.01816} {Visual coreference
  resolution in visual dialog using neural module networks}.

\bibitem[{Kruijff et~al.(2007)Kruijff, Zender, Jensfelt, and
  Christensen}]{Kruijff:2007}
Geert-Jan~M. Kruijff, Hendrik Zender, Patric Jensfelt, and Henrik~I.
  Christensen. 2007.
\newblock Situated dialogue and spatial organization: what, where... and why?
\newblock \emph{International Journal of Advanced Robotic Systems},
  4(1):125--138.

\bibitem[{Kutuzov et~al.(2018)Kutuzov, {\O}vrelid, Szymanski, and
  Velldal}]{Kutuzov2018}
Andrey Kutuzov, Lilja {\O}vrelid, Terrence Szymanski, and Erik Velldal. 2018.
\newblock Diachronic word embeddings and semantic shifts: A survey.
\newblock In \emph{Proceedings of the 27th {{International Conference}} on
  {{Computational Linguistics}}}, pages 1384--1397, {Santa Fe, New Mexico,
  USA}. {Association for Computational Linguistics}.

\bibitem[{Lake et~al.(2019)Lake, Linzen, and Baroni}]{Lake2019}
Brenden~M. Lake, Tal Linzen, and Marco Baroni. 2019.
\newblock \href {http://arxiv.org/abs/1901.04587} {Human few-shot learning of
  compositional instructions}.
\newblock \emph{arXiv:1901.04587 [cs]}.

\bibitem[{Lake et~al.(2017)Lake, Ullman, Tenenbaum, and Gershman}]{Lake:2017ab}
Brenden~M. Lake, Tomer~D. Ullman, Joshua~B. Tenenbaum, and Samuel~J. Gershman.
  2017.
\newblock \href {https://doi.org/10.1017/S0140525X16001837} {Building machines
  that learn and think like people}.
\newblock \emph{Behavioral and Brain Sciences}, 40:e253.

\bibitem[{Lample et~al.(2018)Lample, Conneau, Ranzato, Denoyer, and
  J{\'{e}}gou}]{DBLP:conf/iclr/LampleCRDJ18}
Guillaume Lample, Alexis Conneau, Marc'Aurelio Ranzato, Ludovic Denoyer, and
  Herv{\'{e}} J{\'{e}}gou. 2018.
\newblock \href {https://openreview.net/forum?id=H196sainb} {Word translation
  without parallel data}.
\newblock In \emph{6th International Conference on Learning Representations,
  {ICLR} 2018, Vancouver, BC, Canada, April 30 - May 3, 2018, Conference Track
  Proceedings}. OpenReview.net.

\bibitem[{Larsson(2002)}]{Larsson2002}
Staffan Larsson. 2002.
\newblock \emph{Issue-based Dialogue Management}.
\newblock Ph.D. thesis, University of Gothenburg.

\bibitem[{Larsson(2013)}]{Larsson2013}
Staffan Larsson. 2013.
\newblock \href {https://doi.org/10.1093/logcom/ext059} {Formal semantics for
  perceptual classification}.
\newblock \emph{{Journal of Logic and Computation}}, 25(2):335--369.

\bibitem[{Larsson and Myrendal(2017)}]{Larsson2017}
Staffan Larsson and Jenny Myrendal. 2017.
\newblock \href {https://doi.org/10.21437/SemDial.2017-6} {Dialogue {{Acts}}
  and {{Updates}} for {{Semantic Coordination}}}.
\newblock In \emph{{{SEMDIAL}} 2017 ({{SaarDial}}) {{Workshop}} on the
  {{Semantics}} and {{Pragmatics}} of {{Dialogue}}}, pages 52--59. {ISCA}.

\bibitem[{Lau and Baldwin(2016)}]{DBLP:conf/rep4nlp/LauB16}
Jey~Han Lau and Timothy Baldwin. 2016.
\newblock \href {https://doi.org/10.18653/v1/W16-1609} {An empirical evaluation
  of doc2vec with practical insights into document embedding generation}.
\newblock In \emph{Proceedings of the 1st Workshop on Representation Learning
  for NLP, Rep4NLP@ACL 2016, Berlin, Germany, August 11, 2016}, pages 78--86.
  Association for Computational Linguistics.

\bibitem[{Lavie et~al.(2004)Lavie, Hirst, de~Fockert, and
  Viding}]{Lavie:2004aa}
Nilli Lavie, Aleksandra Hirst, Jan~W de~Fockert, and Essi Viding. 2004.
\newblock \href {https://doi.org/10.1037/0096-3445.133.3.339} {Load theory of
  selective attention and cognitive control}.
\newblock \emph{Journal of Experimental Psychology: General}, 133(3):339--354.

\bibitem[{Lowe(1999)}]{Lowe:1999aa}
David~G Lowe. 1999.
\newblock \href {https://doi.org/10.1109/ICCV.1999.790410} {Object recognition
  from local scale-invariant features}.
\newblock In \emph{Computer vision, 1999. The proceedings of the seventh IEEE
  international conference on}, volume~2, pages 1150--1157. IEEE.

\bibitem[{{Lu} et~al.(2017){Lu}, {Xiong}, {Parikh}, and
  {Socher}}]{lu2017knowing}
J.~{Lu}, C.~{Xiong}, D.~{Parikh}, and R.~{Socher}. 2017.
\newblock \href {https://doi.org/10.1109/CVPR.2017.345} {Knowing when to look:
  Adaptive attention via a visual sentinel for image captioning}.
\newblock In \emph{2017 IEEE Conference on Computer Vision and Pattern
  Recognition (CVPR)}, pages 3242--3250.

\bibitem[{Lu et~al.(2019)Lu, Batra, Parikh, and Lee}]{lu2019vilbert}
Jiasen Lu, Dhruv Batra, Devi Parikh, and Stefan Lee. 2019.
\newblock \href
  {https://proceedings.neurips.cc/paper/2019/file/c74d97b01eae257e44aa9d5bade97baf-Paper.pdf}
  {Vilbert: Pretraining task-agnostic visiolinguistic representations for
  vision-and-language tasks}.
\newblock In \emph{Advances in Neural Information Processing Systems},
  volume~32. Curran Associates, Inc.

\bibitem[{Lu et~al.(2016)Lu, Yang, Batra, and Parikh}]{lu2017hierarchical}
Jiasen Lu, Jianwei Yang, Dhruv Batra, and Devi Parikh. 2016.
\newblock Hierarchical question-image co-attention for visual question
  answering.
\newblock In \emph{Proceedings of the 30th International Conference on Neural
  Information Processing Systems}, NIPS'16, page 289–297, Red Hook, NY, USA.
  Curran Associates Inc.

\bibitem[{{Lücking}(2016)}]{lucking2016}
A.~{Lücking}. 2016.
\newblock Modeling co-verbal gesture perception in type theory with records.
\newblock In \emph{2016 Federated Conference on Computer Science and
  Information Systems (FedCSIS)}, pages 383--392.

\bibitem[{Maraev et~al.(2018)Maraev, Mazzocconi, Howes, and
  Ginzburg}]{Maraev.Mazzocconi.Howes.Ginzburg_ISCA_2018}
Vladislav Maraev, Chiara Mazzocconi, Christine Howes, and Jonathan Ginzburg.
  2018.
\newblock \href {https://doi.org/10.21437/AI-MHRI.2018-3} {{I}ntegrating
  laughter into spoken dialogue systems: preliminary analysis and suggested
  programme}.
\newblock In \emph{Proceedings of the FAIM/ISCA Workshop on Artificial
  Intelligence for Multimodal Human Robot Interaction}, pages 9--14.

\bibitem[{Marsella et~al.(2010)Marsella, Gratch, Petta
  et~al.}]{marsella2010computational}
Stacy Marsella, Jonathan Gratch, Paolo Petta, et~al. 2010.
\newblock Computational models of emotion.
\newblock \emph{A Blueprint for Affective Computing-A sourcebook and manual},
  11(1):21--46.

\bibitem[{Martarelli et~al.(2017)Martarelli, Chiquet, Laeng, and
  Mast}]{MartarelliFred17}
Corinna Martarelli, Sandra Chiquet, Bruno Laeng, and Fred Mast. 2017.
\newblock \href {https://doi.org/10.1007/s00426-016-0781-2} {Using space to
  represent categories: insights from gaze position}.
\newblock \emph{Psychological Research}, 81.

\bibitem[{Mazzocconi(2019)}]{mazzocconi2019phd}
Chiara Mazzocconi. 2019.
\newblock \emph{Laughter in interaction: semantics, pragmatics and child
  development}.
\newblock Ph.D. thesis, Universit{\'e} de Paris.

\bibitem[{Mehu(2011)}]{mehu2011smiling}
Marc Mehu. 2011.
\newblock Smiling and laughter in naturally occurring dyadic interactions:
  Relationship to conversation, body contacts, and displacement activities.
\newblock \emph{Human Ethology Bulletin}, 26(1):10--28.

\bibitem[{Mikolov et~al.(2013{\natexlab{a}})Mikolov, Chen, Corrado, and
  Dean}]{DBLP:journals/corr/abs-1301-3781}
Tom{\'{a}}s Mikolov, Kai Chen, Greg Corrado, and Jeffrey Dean.
  2013{\natexlab{a}}.
\newblock \href {http://arxiv.org/abs/1301.3781} {Efficient estimation of word
  representations in vector space}.
\newblock In \emph{1st International Conference on Learning Representations,
  {ICLR} 2013, Scottsdale, Arizona, USA, May 2-4, 2013, Workshop Track
  Proceedings}.

\bibitem[{Mikolov et~al.(2013{\natexlab{b}})Mikolov, Sutskever, Chen, Corrado,
  and Dean}]{DBLP:conf/nips/MikolovSCCD13}
Tom{\'{a}}s Mikolov, Ilya Sutskever, Kai Chen, Gregory~S. Corrado, and Jeffrey
  Dean. 2013{\natexlab{b}}.
\newblock \href
  {https://proceedings.neurips.cc/paper/2013/hash/9aa42b31882ec039965f3c4923ce901b-Abstract.html}
  {Distributed representations of words and phrases and their
  compositionality}.
\newblock In \emph{Advances in Neural Information Processing Systems 26: 27th
  Annual Conference on Neural Information Processing Systems 2013. Proceedings
  of a meeting held December 5-8, 2013, Lake Tahoe, Nevada, United States},
  pages 3111--3119.

\bibitem[{Mills and Healey(2008)}]{Mills2008}
Gregory Mills and Pat Healey. 2008.
\newblock Semantic negotiation in dialogue: The mechanisms of alignment.
\newblock \emph{Proceedings of the 9th SIGdial Workshop on Discourse and
  Dialogue}, pages 46--53.

\bibitem[{Mishra and Bhattacharyya(2018)}]{Mishra18}
Abhijit Mishra and Pushpak Bhattacharyya. 2018.
\newblock \href {https://doi.org/10.1007/978-981-13-1516-9_4} {\emph{Scanpath
  Complexity: Modeling Reading/Annotation Effort Using Gaze Information: An
  Investigation Based on Eye-tracking}}, pages 77--98. Springer, Singapore.

\bibitem[{Mitchell et~al.(2012)Mitchell, Han, Dodge, Mensch, Goyal, Berg,
  Yamaguchi, Berg, Stratos, and Daum{\'e}~III}]{Mitchell:2012aa}
Margaret Mitchell, Xufeng Han, Jesse Dodge, Alyssa Mensch, Amit Goyal, Alex
  Berg, Kota Yamaguchi, Tamara Berg, Karl Stratos, and Hal Daum{\'e}~III. 2012.
\newblock Midge: Generating image descriptions from computer vision detections.
\newblock In \emph{Proceedings of the 13th Conference of the European Chapter
  of the Association for Computational Linguistics}, pages 747--756.
  Association for Computational Linguistics.

\bibitem[{Montague(1973)}]{Montague1973}
Richard Montague. 1973.
\newblock {The Proper Treatment of Quantification in Ordinary English}.
\newblock In Jaakko Hintikka, Julius Moravcsik, and Patrick Suppes, editors,
  \emph{{Approaches to Natural Language: Proceedings of the 1970 Stanford
  Workshop on Grammar and Semantics}}, pages 247--270. D. Reidel Publishing
  Company, Dordrecht.

\bibitem[{Mooney(2008)}]{mooney2008}
Raymond~J. Mooney. 2008.
\newblock Learning to connect language and perception.
\newblock In \emph{Proceedings of the 23rd National Conference on Artificial
  Intelligence - Volume 3}, AAAI'08, page 1598–1601. AAAI Press.

\bibitem[{Morett et~al.(2020)Morett, Hughes~Berheim, and Bulger}]{Morett20}
Laura Morett, Sarah Hughes~Berheim, and Raymond Bulger. 2020.
\newblock \href {https://doi.org/10.3389/fpsyg.2020.575991} {Semantic
  relationships between representational gestures and their lexical affiliates
  are evaluated similarly for speech and text}.
\newblock \emph{Frontiers in Psychology}, 11.

\bibitem[{Myrendal(2015)}]{Myrendal2015}
Jenny Myrendal. 2015.
\newblock \emph{Word {{Meaning Negotiation}} in {{Online Discussion Forum
  Communication}}}.
\newblock {{PhD Thesis}}, University of Gothenburg, {University of Gothenburg}.

\bibitem[{Nor{\'e}n and Linell(2007)}]{Noren2007}
Kerstin Nor{\'e}n and Per Linell. 2007.
\newblock \href {https://doi.org/10.1075/prag.17.3.03nor} {Meaning potentials
  and the interaction between lexis and contexts: {{An}} empirical
  substantiation}.
\newblock \emph{Pragmatics}, 17(3):387--416.

\bibitem[{Oatley and Johnson-Laird(2014)}]{oatley14}
Keith Oatley and P.N. Johnson-Laird. 2014.
\newblock \href {https://doi.org/10.1016/j.tics.2013.12.004} {Cognitive
  approaches to emotions}.
\newblock \emph{Trends in Cognitive Sciences}, 18(3):134--140.

\bibitem[{{\"O}stling and Tiedemann(2017)}]{ostling-tiedemann-2017-continuous}
Robert {\"O}stling and J{\"o}rg Tiedemann. 2017.
\newblock \href {https://www.aclweb.org/anthology/E17-2102} {Continuous
  multilinguality with language vectors}.
\newblock In \emph{Proceedings of the 15th Conference of the {E}uropean Chapter
  of the Association for Computational Linguistics: Volume 2, Short Papers},
  pages 644--649, Valencia, Spain. Association for Computational Linguistics.

\bibitem[{Partee(1976)}]{Partee1976}
Barbara~H. Partee, editor. 1976.
\newblock \emph{{Montague Grammar}}.
\newblock Academic Press.

\bibitem[{Paul(1891)}]{Paul1891}
Hermann Paul. 1891.
\newblock \emph{Principles of the History of Language}.
\newblock {London ; New York : Longmans, Green}.

\bibitem[{Pires et~al.(2019)Pires, Schlinger, and
  Garrette}]{pires-etal-2019-multilingual}
Telmo Pires, Eva Schlinger, and Dan Garrette. 2019.
\newblock \href {https://doi.org/10.18653/v1/P19-1493} {How multilingual is
  multilingual {BERT}?}
\newblock In \emph{Proceedings of the 57th Annual Meeting of the Association
  for Computational Linguistics}, pages 4996--5001, Florence, Italy.
  Association for Computational Linguistics.

\bibitem[{Prusak(2006)}]{prusak2006science}
Bernard~G Prusak. 2006.
\newblock The science of laughter: Helmuth plessner’s laughing and crying
  revisited.
\newblock \emph{Continental philosophy review}, 38:41--69.

\bibitem[{Ramisa et~al.(2015)Ramisa, Wang, Lu, Dellandrea, Moreno-Noguer, and
  Gaizauskas}]{Ramisa:2015aa}
Arnau Ramisa, Josiah Wang, Ying Lu, Emmanuel Dellandrea, Francesc
  Moreno-Noguer, and Robert Gaizauskas. 2015.
\newblock \href {https://www.aclweb.org/anthology/D15-1022/} {Combining
  geometric, textual and visual features for predicting prepositions in image
  descriptions}.
\newblock In \emph{Proceedings of the 2015 Conference on Empirical Methods in
  Natural Language Processing}, pages 214--220, Lisbon, Portugal. Association
  for Computational Linguistics.

\bibitem[{Regier(1996)}]{regier1996}
Terry Regier. 1996.
\newblock \emph{The human semantic potential spatial language and constrained
  connectionism}.
\newblock Neural network modeling and connectionism. MIT Press, Cambridge.

\bibitem[{Reyle(1993)}]{Reyle1993}
Uwe Reyle. 1993.
\newblock {Dealing with ambiguities by underspecification: Construction,
  representation and deduction}.
\newblock \emph{Journal of Semantics}, 10(2):123--179.

\bibitem[{Rosenfeld and Erk(2018)}]{Rosenfeld2018}
Alex Rosenfeld and Katrin Erk. 2018.
\newblock \href {https://doi.org/10.18653/v1/N18-1044} {Deep {{Neural Models}}
  of {{Semantic Shift}}}.
\newblock In \emph{Proceedings of the 2018 {{Conference}} of the {{North
  American Chapter}} of the {{Association}} for {{Computational Linguistics}}:
  {{Human Language}} {{Technologies}}, {{Volume}} 1 ({{Long Papers}})}, pages
  474--484, {New Orleans, Louisiana}. {Association for Computational
  Linguistics}.

\bibitem[{Sadeghi et~al.(2015)Sadeghi, Kumar~Divvala, and
  Farhadi}]{Sadeghi:2015aa}
Fereshteh Sadeghi, Santosh~K Kumar~Divvala, and Ali Farhadi. 2015.
\newblock \href
  {https://www.cv-foundation.org/openaccess/content_cvpr_2015/html/Sadeghi_VisKE_Visual_Knowledge_2015_CVPR_paper.html}
  {Viske: Visual knowledge extraction and question answering by visual
  verification of relation phrases}.
\newblock In \emph{Proceedings of the IEEE conference on computer vision and
  pattern recognition}, pages 1456--1464.

\bibitem[{Scherer(2009)}]{scherer2009dynamic}
Klaus~R Scherer. 2009.
\newblock The dynamic architecture of emotion: Evidence for the component
  process model.
\newblock \emph{Cognition and emotion}, 23(7):1307--1351.

\bibitem[{Schlangen(2019)}]{schlangen2019grounded}
David Schlangen. 2019.
\newblock \href {http://arxiv.org/abs/1908.11279} {Grounded agreement games:
  Emphasizing conversational grounding in visual dialogue settings}.

\bibitem[{Somashekarappa et~al.(2020)Somashekarappa, Howes, and
  Sayeed}]{Vidya20}
Vidya Somashekarappa, Christine Howes, and Asad Sayeed. 2020.
\newblock \href {https://www.aclweb.org/anthology/2020.lrec-1.95} {An
  annotation approach for social and referential gaze in dialogue}.
\newblock In \emph{Proceedings of the 12th Language Resources and Evaluation
  Conference}, pages 759--765, Marseille, France. European Language Resources
  Association.

\bibitem[{Spivey et~al.(2000)Spivey, Richardson, Tyler, and Young}]{Spivey00}
Michael Spivey, Daniel Richardson, Melinda Tyler, and Ezekiel~E Young. 2000.
\newblock Eye movements during comprehension of spoken descriptions.
\newblock In \emph{Proceedings of the 22nd Annual Meeting of the Cognitive
  Science Society}.

\bibitem[{Stalnaker(2002)}]{Stalnaker2002}
Robert Stalnaker. 2002.
\newblock Common {{Ground}}.
\newblock \emph{Linguistics and Philosophy}, 25(5-6):701--721.

\bibitem[{Su et~al.(2020)Su, Zhu, Cao, Li, Lu, Wei, and Dai}]{Su2019}
Weijie Su, Xizhou Zhu, Yue Cao, Bin Li, Lewei Lu, Furu Wei, and Jifeng Dai.
  2020.
\newblock \href {https://openreview.net/forum?id=SygXPaEYvH} {Vl-bert:
  Pre-training of generic visual-linguistic representations}.
\newblock In \emph{International Conference on Learning Representations}.

\bibitem[{Tahmasebi et~al.(2018)Tahmasebi, Borin, and Jatowt}]{Tahmasebi2018}
Nina Tahmasebi, Lars Borin, and Adam Jatowt. 2018.
\newblock \href {http://arxiv.org/abs/1811.06278} {Survey of {{Computational
  Approaches}} to {{Diachronic Conceptual Change}}}.
\newblock \emph{arXiv:1811.06278 [cs]}.

\bibitem[{Tang(2018)}]{Tang2018}
Xuri Tang. 2018.
\newblock \href {https://doi.org/10.1017/S1351324918000220} {A state-of-the-art
  of semantic change computation}.
\newblock \emph{Natural Language Engineering}, 24(5):649--676.

\bibitem[{Tellex et~al.(2011)Tellex, Kollar, Dickerson, Walter, Banerjee,
  Teller, and Roy}]{Tellex:2011wf}
Stefanie Tellex, Thomas Kollar, Steven Dickerson, Matthew Walter, Ashis
  Banerjee, Seth Teller, and Nicholas Roy. 2011.
\newblock \href {https://ojs.aaai.org/index.php/AAAI/article/view/7979}
  {Understanding natural language commands for robotic navigation and mobile
  manipulation}.
\newblock \emph{Proceedings of the AAAI Conference on Artificial Intelligence},
  25(1).

\bibitem[{Tenenbaum(2020)}]{Tenenbaum:2020aa}
Joshua~B. Tenenbaum. 2020.
\newblock \href
  {https://slideslive.com/38929461/cognitive-and-computational-building-blocks-for-more-humanlike-language-in-machines}
  {Cognitive and computational building blocks for morehuman-like language in
  machines}.
\newblock Acl 2020 keynote, Center for Brains, Minds and Machines, MIT.

\bibitem[{Vaswani et~al.(2017)Vaswani, Shazeer, Parmar, Uszkoreit, Jones,
  Gomez, Kaiser, and Polosukhin}]{DBLP:conf/nips/VaswaniSPUJGKP17}
Ashish Vaswani, Noam Shazeer, Niki Parmar, Jakob Uszkoreit, Llion Jones,
  Aidan~N. Gomez, Lukasz Kaiser, and Illia Polosukhin. 2017.
\newblock \href
  {https://proceedings.neurips.cc/paper/2017/hash/3f5ee243547dee91fbd053c1c4a845aa-Abstract.html}
  {Attention is all you need}.
\newblock In \emph{Advances in Neural Information Processing Systems 30: Annual
  Conference on Neural Information Processing Systems 2017, December 4-9, 2017,
  Long Beach, CA, {USA}}, pages 5998--6008.

\bibitem[{Vinyals et~al.(2015)Vinyals, Toshev, Bengio, and
  Erhan}]{vinyals2015tell}
Oriol Vinyals, Alexander Toshev, Samy Bengio, and Dumitru Erhan. 2015.
\newblock \href {http://arxiv.org/abs/1411.4555} {Show and tell: A neural image
  caption generator}.

\bibitem[{de~Vries et~al.(2017)de~Vries, Strub, Chandar, Pietquin, Larochelle,
  and Courville}]{devries2017guesswhat}
Harm de~Vries, Florian Strub, Sarath Chandar, Olivier Pietquin, Hugo
  Larochelle, and Aaron Courville. 2017.
\newblock \href {http://arxiv.org/abs/1611.08481} {Guesswhat?! visual object
  discovery through multi-modal dialogue}.

\bibitem[{Vulic et~al.(2020)Vulic, Baker, Ponti, Petti, Leviant, Wing,
  Majewska, Bar, Malone, Poibeau, Reichart, and
  Korhonen}]{DBLP:journals/corr/abs-2003-04866}
Ivan Vulic, Simon Baker, Edoardo~Maria Ponti, Ulla Petti, Ira Leviant, Kelly
  Wing, Olga Majewska, Eden Bar, Matt Malone, Thierry Poibeau, Roi Reichart,
  and Anna Korhonen. 2020.
\newblock \href {http://arxiv.org/abs/2003.04866} {Multi-simlex: {A}
  large-scale evaluation of multilingual and cross-lingual lexical semantic
  similarity}.
\newblock \emph{CoRR}, abs/2003.04866.

\bibitem[{Wilkins(2006)}]{David06}
David Wilkins. 2006.
\newblock \href {https://doi.org/10.1075/gest.6.1.08wil} {Adam kendon (2004).
  gesture: Visible action as utterance}.
\newblock \emph{Gesture}, 6.

\bibitem[{Willems et~al.(2007)Willems, Özyürek, and Hagoort}]{Willems07}
Roel Willems, Asli Özyürek, and Peter Hagoort. 2007.
\newblock \href {https://doi.org/10.1093/cercor/bhl141} {When language meets
  action: The neural integration of gesture and speech}.
\newblock \emph{Cerebral cortex (New York, N.Y. : 1991)}, 17:2322--33.

\bibitem[{Wittgenstein(1953)}]{wittgenstein1953philosophical}
Ludwig Wittgenstein. 1953.
\newblock \emph{Philosophical Investigations}.
\newblock Basil Blackwell, Oxford.

\bibitem[{Xu et~al.(2015)Xu, Ba, Kiros, Cho, Courville, Salakhudinov, Zemel,
  and Bengio}]{xu2016show}
Kelvin Xu, Jimmy Ba, Ryan Kiros, Kyunghyun Cho, Aaron Courville, Ruslan
  Salakhudinov, Rich Zemel, and Yoshua Bengio. 2015.
\newblock \href {http://proceedings.mlr.press/v37/xuc15.html} {Show, attend and
  tell: Neural image caption generation with visual attention}.
\newblock In \emph{Proceedings of the 32nd International Conference on Machine
  Learning}, volume~37 of \emph{Proceedings of Machine Learning Research},
  pages 2048--2057, Lille, France. PMLR.

\end{thebibliography}
\bibliographystyle{acl_natbib}

\end{document}